\documentclass[11pt]{article}
\usepackage[margin=1in]{geometry}
\usepackage[numbers]{natbib}
\usepackage{times}
\usepackage{latexsym}
\usepackage[T1]{fontenc}
\usepackage[utf8]{inputenc}
\usepackage{microtype}
\usepackage{graphicx}
\usepackage{booktabs}
\usepackage{amsmath}
\usepackage{multirow}
\usepackage{url}

\title{Bigger or Cheaper? Scale and Quantization Effects on\\
       Uncertainty Signals in Vision-Language Models\\
       Under Image Degradation}

\author{M M Asif Ferdous \\
        Independent Researcher \\
        \texttt{ferdous.asif2012@gmail.com}}

\begin{document}
\maketitle

% ---------------------------------------------------------------------------
\begin{abstract}
Vision-language models (VLMs) deployed on consumer hardware must decide when to
answer and when to defer, and that decision depends on having a confidence
signal that tracks correctness. A practitioner with a fixed memory budget faces
a choice between a small model at full precision, the same small model
quantized, and a larger model quantized into the same footprint --- three
configurations that push the confidence signal in opposing directions.
We measure, on identical inputs, how model scale and 4-bit quantization affect
two confidence signals in the Qwen2-VL family: the confidence a model
\emph{states} in natural language, and its own mean token probability over the
answer it generates. Across 5{,}700 predictions spanning six realistic
photographic degradations at three severities, we find that scale sharply
improves the model's \emph{internal} uncertainty signal (mean error-detection
AUROC $0.80\!\to\!0.98$ from 2B to 7B) while its \emph{verbalized} confidence
stays weak and often at chance (mean $0.61\!\to\!0.69$): the gap between what the
model knows and what it says widens rather than closes with size. We find that
4-bit quantization is nearly free for accuracy ($-1.6$ points) but expensive for
the confidence signal (internal AUROC $0.95\!\to\!0.80$, and the
verbalized-confidence parse rate collapses from $99\%$ to $64\%$). For a fixed
memory budget the recommendation is therefore to prefer a larger quantized model
over a smaller full-precision one: 7B-4bit gives both the best accuracy and the
best uncertainty signal (internal AUROC $0.98$) of the three configurations that
fit. We frame the results as selective-prediction operating points so they
translate directly into a deployment recommendation, and we argue that
error-detection AUROC, not calibration error, is the metric that exposes the
difference between the two signals.
\end{abstract}

% ===========================================================================
\section{Introduction}
% ===========================================================================
Vision-language models are increasingly deployed at the small end of the
parameter range, on consumer GPUs and edge devices chosen for cost, latency,
privacy, or the absence of reliable connectivity. In these settings the images
reaching the model are rarely clean benchmark photographs: they are compressed
by messaging applications, blurred by handheld capture, underexposed indoors, or
washed out by direct light.

In such deployments a usable uncertainty signal matters as much as raw accuracy,
because it determines whether the system can \emph{defer}. A model that is wrong
and signals doubt can hand off to a human; a model that is wrong and signals
confidence cannot. Generative VLMs expose two routes to such a signal. The first
is to prompt the model to state its confidence as a number --- \emph{verbalized
confidence}. The second is to read the model's own probability over the tokens
it generated --- \emph{internal confidence}. These are not obviously the same
quantity, and how each behaves as a function of the two levers a practitioner
actually controls --- model size and numerical precision --- has not been
measured for open-weight VLMs under image degradation.

The practical form of the problem is a budget constraint. Consider a fixed
16\,GB GPU. At least three configurations fit:

\begin{itemize}
  \item a \textbf{small model at full precision} (2B, fp16);
  \item the \textbf{same small model quantized} (2B, 4-bit), the cheapest and
        fastest option;
  \item a \textbf{larger model quantized} into the same footprint (7B, 4-bit),
        the most capable option numerically.
\end{itemize}

\noindent All three run within budget, and the practitioner must choose one. The
two axes pull against each other: more parameters should improve the model's
self-knowledge, while lower precision should perturb exactly the probability
distribution from which internal confidence is read. The net effect on the
reliability of the confidence signal is unknown.

This paper measures it. We ask:

\begin{itemize}
  \item[\textbf{RQ1}] Does model scale improve the quality of a VLM's
        uncertainty signals under image degradation?
  \item[\textbf{RQ2}] Does 4-bit quantization degrade those signals, and by how
        much?
  \item[\textbf{RQ3}] Under a fixed memory budget, which configuration should a
        practitioner deploy?
\end{itemize}

We answer them with a three-arm study on the Qwen2-VL family, run on identical
images and prompts: 2B at fp16, 2B at 4-bit, and 7B at 4-bit. Holding precision
fixed and varying parameters isolates the \emph{scale} effect (RQ1); holding
parameters fixed and varying precision isolates the \emph{quantization} effect
(RQ2). We evaluate both confidence signals on every prediction, using
error-detection AUROC --- the probability that a correct answer is assigned
higher confidence than an incorrect one --- as the primary metric, because it
measures directly whether a signal is usable for deferral.

Our contributions are:
\begin{enumerate}
  \item A controlled measurement of how \textbf{model scale} affects verbalized
        and internal confidence in VLMs under six realistic degradations, with
        bootstrap confidence intervals.
  \item A controlled measurement of how \textbf{4-bit quantization} affects the
        same signals, with model size held constant --- a confound that would
        otherwise contaminate any scale comparison drawn across precisions.
  \item A \textbf{deployment recommendation} expressed as
        selective-prediction operating points (risk--coverage curves, area under
        the risk--coverage curve, and threshold-transfer behaviour under
        degradation), turning the measurements into an actionable choice.
  \item A methodological argument, supported by our data, that
        \textbf{error-detection AUROC rather than calibration error} is the
        diagnostic metric for confidence signals whose variance is near zero.
\end{enumerate}

% ===========================================================================
\section{Related Work}
% ===========================================================================

\paragraph{Verbalized uncertainty in language and vision-language models.}
Because generative models emit text rather than a probability vector, a natural
route to uncertainty is to prompt the model to state its confidence.
\citet{xiong2024llms} evaluate confidence-elicitation strategies in LLMs and
report systematic overconfidence, with most stated scores clustering in the
80--100 range. \citet{tian2023just} find that for RLHF-tuned models verbalized
confidence can be better calibrated than sampling-based estimates, a result that
made verbalized uncertainty a common default. \citet{groot2024overconfidence}
extend verbalized confidence estimation to visual question answering and find
VLMs poorly calibrated and severely overconfident. Our study measures verbalized
confidence as one of two signals, but does so across a scale and precision grid
rather than at a single operating point.

\paragraph{Calibration and model scale.}
Whether stated confidence improves with scale is contested. Across LLM families,
calibration tends to improve with size, but a recurring finding is that most of
that improvement comes from rising accuracy rather than from reduced
overconfidence, and overconfidence persists at every size tested
\citep{xiong2024llms}. Whether this pattern transfers to the vision-language
setting, where the confidence must be conditioned on a degraded image, is
untested. RQ1 addresses exactly this gap.

\paragraph{Quantization and confidence.}
Post-training quantization is standard practice for fitting larger models onto
constrained hardware, but it perturbs the output distribution. Empirical studies
report that 4-bit models tend to show slightly lower confidence than their
full-precision counterparts, with larger drops on some model/dataset pairs
\citep{quant2025interpreting}. Because internal confidence is read directly off
the token distribution, any comparison that varies precision and scale together
confounds the two. Our design isolates the quantization effect with a
precision-only contrast at fixed model size (RQ2); to our knowledge this effect
has not been measured for VLM error-detection specifically.

\paragraph{Verbalized uncertainty under image corruption.}
Closest in setting is \citet{borszukovszki2025know}, who evaluate proprietary
VLMs on synthetic corruptions at graded severities and report rising calibration
error and persistent overconfidence, and who identify the inaccessibility of
internal token probabilities as a defining constraint of the proprietary setting.
We work with open-weight models, which makes the internal signal available, and
we vary scale and precision rather than corruption family alone. We follow the
methodological template of graded-severity corruption established by
\citet{hendrycks2019benchmarking} and extended by \citet{michaelis2019benchmarking},
and report metrics as a function of severity rather than in aggregate.

\paragraph{Selective prediction and calibration metrics.}
Selective prediction --- abstaining on low-confidence inputs to raise accuracy on
the rest --- is the deployment framing for a confidence signal, summarised by
risk--coverage curves and the area beneath them. Separately,
\citet{guo2017calibration} showed that post-hoc temperature scaling corrects
classifier overconfidence by redistributing probability mass, and
\citet{ovadia2019trust} established that predictive uncertainty degrades under
dataset shift. We connect these threads: we report calibration error for
comparability but argue, and show on our data, that error-detection AUROC is the
more diagnostic measure when a confidence signal has near-zero variance, since
such a signal can appear well calibrated while carrying no usable ranking
information.

% ===========================================================================
\section{Method}
% ===========================================================================

\subsection{Experimental design}
We run three arms on the Qwen2-VL family \citep{wang2024qwen2}, all on identical
images and prompts:

\begin{center}
\begin{tabular}{@{}clll@{}}
\toprule
Arm & Model & Precision & Role \\
\midrule
A & Qwen2-VL-2B-Instruct & fp16      & small, full precision \\
B & Qwen2-VL-2B-Instruct & NF4 4-bit & small, quantized \\
C & Qwen2-VL-7B-Instruct & NF4 4-bit & large, quantized \\
\bottomrule
\end{tabular}
\end{center}

\noindent This yields two orthogonal contrasts: \textbf{B vs.\ C} isolates the
effect of scale with precision held constant, and \textbf{A vs.\ B} isolates the
effect of quantization with parameters held constant. Using a single model
family throughout ensures the scale contrast is not confounded by differences in
architecture or training recipe across labs. All three arms are generated for
this study; no numbers are imported from external runs.

\subsection{Models and inference}
Both model sizes are open-weight, instruction-tuned Qwen2-VL checkpoints. The
quantized arms use 4-bit NormalFloat (NF4) with double quantization and an
fp16 compute dtype, which fits the 7B model within a single free-tier
16\,GB NVIDIA T4. Decoding is greedy ($\text{do\_sample}=\text{False}$)
throughout for reproducibility, with a 32-token generation cap.

\subsection{Data}
We use a 100-item subset of the Food101 test split \citep{bossard2014food}
(streamed, seed~0), posed as four-option multiple choice: one correct label and
three distractors sampled uniformly from the remaining classes. Food101 provides
real photographs at usable resolution, appropriate for a corruption study. The
same 100 images and the same option sets are used in all three arms; the
pipeline records an item manifest on the first run and aborts any later run that
does not reproduce it exactly, guaranteeing cross-arm comparability.

\subsection{Degradations}
Six degradation families, chosen to approximate real phone-camera artifacts, are
each applied at three severity levels (Table~\ref{tab:deg}). With the clean
baseline this gives 19 conditions per arm and 1{,}900 predictions per
arm ($5{,}700$ across the three arms).

\begin{table}[t]
\centering
\small
\begin{tabular}{@{}lll@{}}
\toprule
Family & Mechanism & Severities 1/2/3 \\
\midrule
JPEG        & re-encode at low quality        & $q=30/15/7$ \\
Motion blur & horizontal box kernel           & $r=2/4/7$ \\
Low light   & brightness $\downarrow$ + noise & $\times0.5/0.3/0.15$ \\
Glare       & brightness amplification        & $\times1.6/2.2/3.0$ \\
Rotation    & in-plane tilt, black fill       & $5^{\circ}/12^{\circ}/20^{\circ}$ \\
Resample    & downscale then upscale          & $\times0.5/0.3/0.15$ \\
\bottomrule
\end{tabular}
\caption{Degradation families and severity parameters.}
\label{tab:deg}
\end{table}

\subsection{Confidence signals}
\paragraph{Verbalized confidence.}
The model is prompted to answer and state an integer confidence in $[0,100]$,
normalised to $[0,1]$. Because verbalized confidence is prompt-sensitive, we do
not fix a template silently: we pilot three templates on 20 clean items per arm,
report the parse rate of each per model size, and use the highest-parse-rate
template for the main runs. The parse-rate comparison across model sizes is
itself reported (Section~\ref{sec:results}). Templates appear verbatim in
Appendix~\ref{app:templates}.

\paragraph{Internal confidence.}
During generation we retain per-step output distributions and compute the mean
probability assigned to each token actually emitted \emph{within the answer
span}:
\begin{equation}
c_{\text{int}} = \frac{1}{T}\sum_{t=1}^{T} p_\theta\!\left(y_t \mid y_{<t}, x\right),
\end{equation}
where $y_t$ is the $t$-th generated token and $x$ the multimodal input.
Restricting the average to the answer span is essential: with a two-line
template the generation contains both the answer and the verbalized
``Confidence:'' digits, and including the latter would fold the verbalized signal
into the internal one, making the central comparison circular. The span is
isolated by mapping character offsets back to token indices. This quantity is
continuous in $(0,1]$ and requires no additional forward passes.

\subsection{Answer matching}
Generated answers rarely match gold labels exactly. We normalise case and
punctuation, accept exact matches, and fall back to unique containment (the
prediction contains exactly one option string). The match method is recorded per
prediction. Crucially, smaller and quantized models frequently ignore the
two-line template and reply with a bare label (``Beignets'') rather than
``Answer: beignets''; such replies are correct and are scored against the whole
reply, so no model is penalised for failing to follow the format. This matters
for a fair cross-model comparison: scoring only an explicit answer field would
understate the accuracy of the arm with the lowest format compliance and confound
capability with instruction-following. Verbalized-confidence parsing is unaffected
--- a bare reply still yields no confidence value --- so the verbalized parse rate
is reported separately as a signal in its own right.

\subsection{Metrics}
We report accuracy, Expected Calibration Error \citep{naeini2015obtaining} with
10 bins, Brier score \citep{brier1950verification}, and \textbf{error-detection
AUROC} --- the probability that a randomly chosen correct answer receives higher
confidence than a randomly chosen incorrect one. AUROC is our primary metric
because it measures directly whether a signal can be thresholded for deferral.
Confidence intervals are percentile bootstrap with $B=2000$ resamples.
Conditions at or near 100\% accuracy contain too few errors for AUROC to be
defined; these are reported as n/a and excluded from aggregates.

\subsection{Selective prediction analysis}
From the same predictions, at no additional inference cost, we compute
risk--coverage curves (accuracy retained as a function of the fraction of inputs
answered), the area under the risk--coverage curve (AURC) per condition and
signal, and a threshold-transfer analysis: a deferral threshold fit on clean
images to answer a fixed fraction of them is applied unchanged to each degraded
condition, and we report where it holds and where it silently admits errors.

% ===========================================================================
\section{Results}
\label{sec:results}
% ===========================================================================
Table~\ref{tab:means} summarises the three arms. Accuracy is high in every arm;
the arms differ far more in the \emph{quality of their confidence signals} than
in raw accuracy, which is the paper's central observation.

\begin{table}[t]
\centering
\small
\begin{tabular}{@{}lccc@{}}
\toprule
 & 2B fp16 & 2B 4-bit & 7B 4-bit \\
 & (A) & (B) & (C) \\
\midrule
Accuracy               & 0.904 & 0.888 & \textbf{0.935} \\
Verbalized parse rate  & 0.99  & 0.64  & \textbf{1.00}  \\
Verbalized AUROC       & 0.56  & 0.61  & 0.69           \\
Internal AUROC         & 0.95  & 0.80  & \textbf{0.98}  \\
Internal AURC $\downarrow$ & 0.017 & 0.048 & \textbf{0.008} \\
\bottomrule
\end{tabular}
\caption{Means over 19 conditions. AUROC is error detection (higher is better);
AURC is area under the risk--coverage curve (lower is better). The large gaps are
in the confidence signals, not in accuracy.}
\label{tab:means}
\end{table}

\subsection{Accuracy is robust; scale helps most where it is hardest}
\begin{figure}[t]
\centering
\includegraphics[width=0.95\columnwidth]{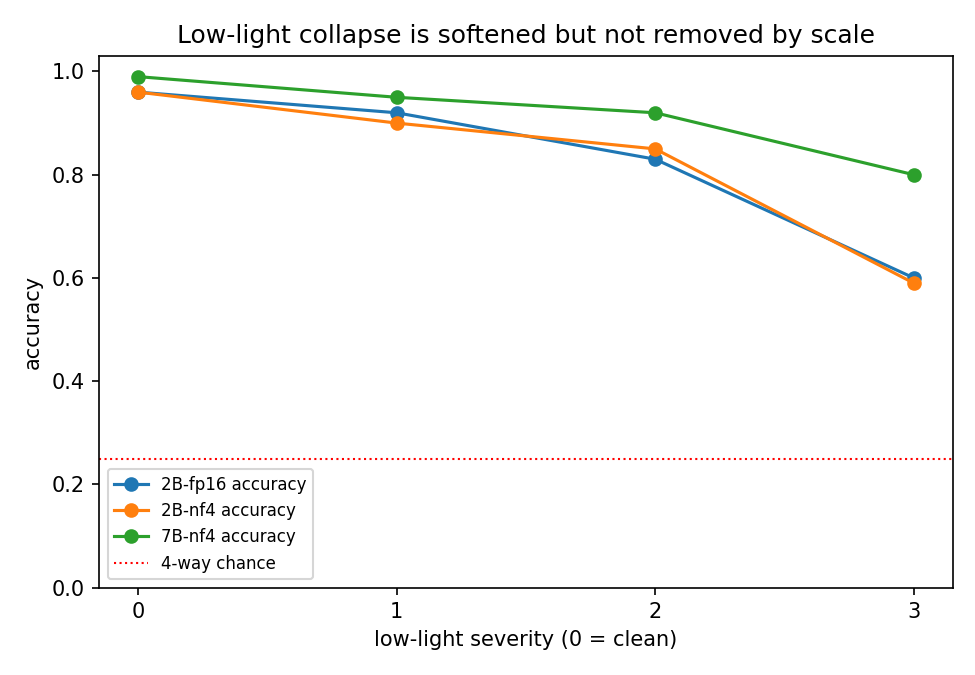}
\caption{Accuracy under increasing low light. Scale softens but does not remove
the collapse; all arms fall far below the clean baseline at severity 3.}
\label{fig:lowlight}
\end{figure}

Accuracy is high and comparable across arms (0.88--0.94 overall). It is nearly
insensitive to JPEG compression, blur, rotation and resampling in all three arms,
and degrades under severe glare and low light. Scale helps precisely on the
hardest conditions: from 2B-4bit to 7B-4bit, accuracy rises by $+0.21$ at low
light s3, $+0.13$ at both glare s3 and JPEG s3, and by $\le0.05$ elsewhere
(Figure~\ref{fig:lowlight}). The low-light collapse is softened but not removed:
low light s3 accuracy is 0.59 (2B-4bit), 0.60 (2B-fp16) and 0.80 (7B-4bit),
all well below the clean baseline.

\subsection{RQ1: scale improves the internal signal far more than the verbalized
one}
Holding precision fixed at 4-bit, increasing scale from 2B to 7B raises mean
internal error-detection AUROC from 0.80 to 0.98, and the 7B internal signal is
$\ge0.90$ in 18 of 19 conditions (range 0.89--1.00). Verbalized confidence
improves far less: mean AUROC rises only from 0.61 to 0.69, and remains at or
near chance ($\le0.60$) in 8 of 19 conditions at 7B, including every low-light
condition (0.51--0.56). The verbalized parse rate does improve sharply with scale
(0.64 to 1.00): the larger model reliably produces a confidence number, it simply
is not a discriminative one.

\begin{figure*}[t]
\centering
\includegraphics[width=0.92\textwidth]{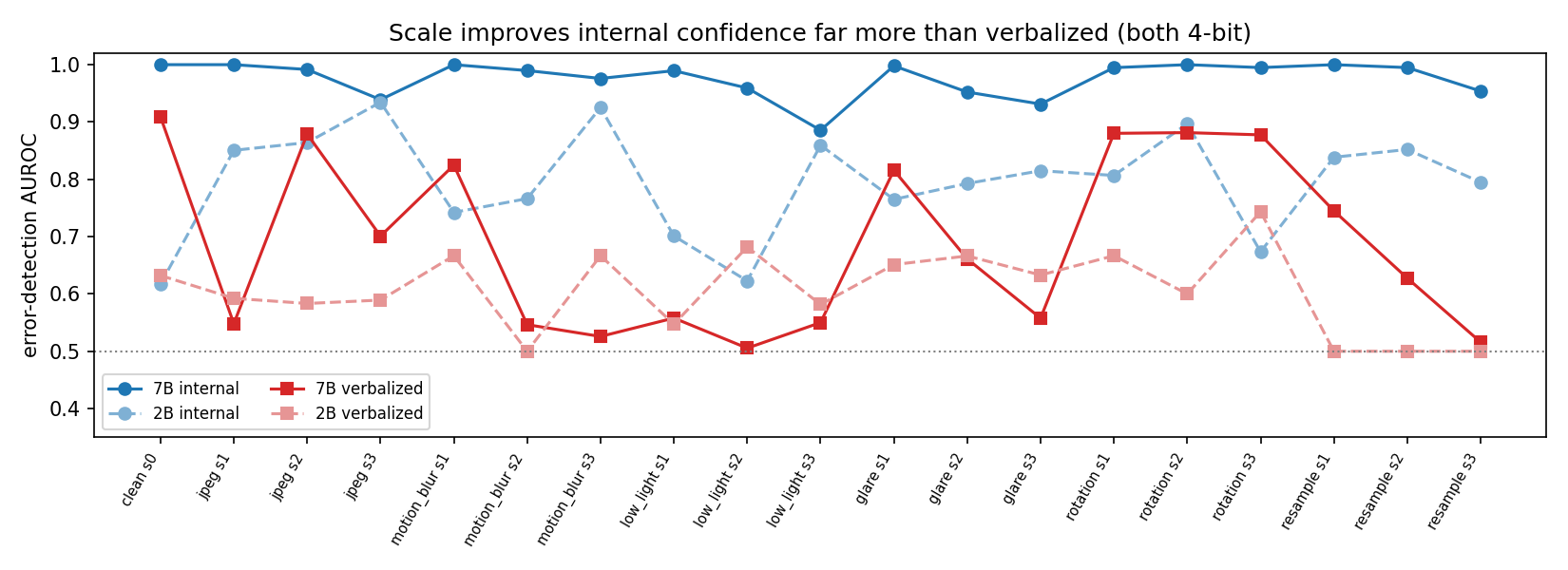}
\caption{Error-detection AUROC per condition, 2B vs 7B (both 4-bit). The 7B
internal signal (dark blue) sits near 1.0 almost everywhere; verbalized confidence
(red) swings around chance. The internal--verbalized gap is larger at 7B than at
2B: scale improves what the model knows about its errors much more than what it
says.}
\label{fig:scale}
\end{figure*}

The consequence, visible in Figure~\ref{fig:scale}, is that the gap between the
model's internal self-knowledge and its verbalized self-report \emph{widens} with
scale rather than closing: 0.19 AUROC at 2B ($0.80$ vs $0.61$) versus 0.29 at 7B
($0.98$ vs $0.69$). Scale makes the model know its own errors much better without
making it say so much better. This is our central finding: verbalized confidence
in these VLMs is not a small-model artifact that scale repairs; the internal
distribution and the verbalized number are governed by different mechanisms, and
only the former tracks correctness.

\subsection{RQ2: quantization is cheap for accuracy, expensive for the signal}
\begin{figure}[t]
\centering
\includegraphics[width=0.95\columnwidth]{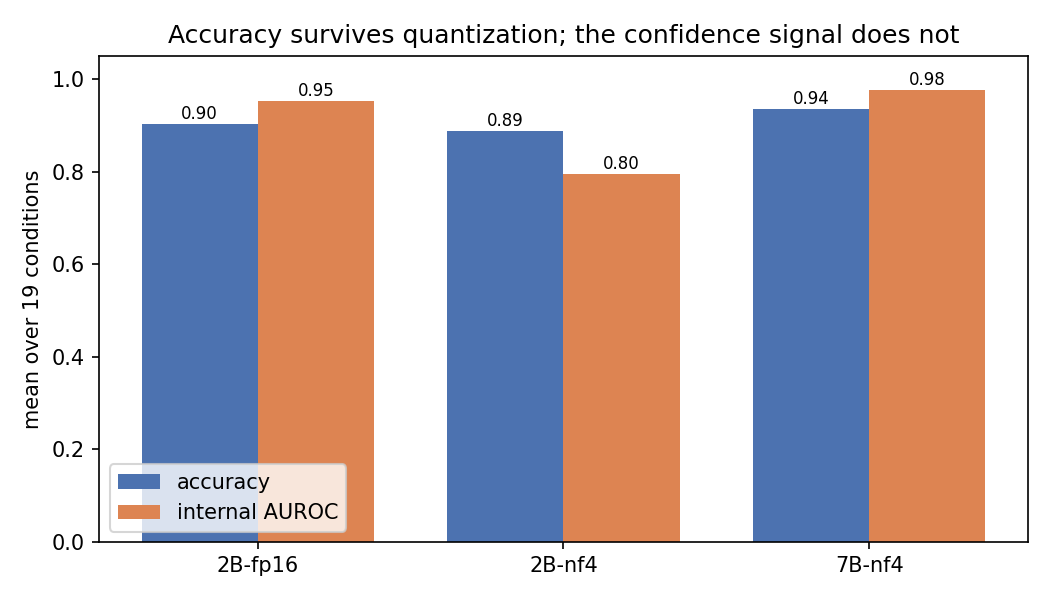}
\caption{Quantization effect at 2B (means over 19 conditions). Accuracy is nearly
unchanged from fp16 to 4-bit, but internal error-detection AUROC drops from 0.95
to 0.80.}
\label{fig:quant}
\end{figure}

Holding scale fixed at 2B, moving from fp16 to 4-bit costs only $1.6$ accuracy
points (0.904 to 0.888), consistent with quantization being an attractive way to
save memory. The confidence signal, however, is hit hard: internal AUROC falls
from 0.95 to 0.80, and the verbalized parse rate collapses from 0.99 to 0.64 ---
the quantized model frequently abandons the requested two-line format and returns
a bare label. Quantization therefore preserves what the model answers while
degrading the model's expressed and internal knowledge of \emph{whether it is
right} (Figure~\ref{fig:quant}). A deployment that quantizes a small VLM to fit
its budget keeps accuracy but loses much of the signal needed to defer safely.

\subsection{RQ3: for a fixed budget, prefer the larger quantized model}
\begin{figure}[t]
\centering
\includegraphics[width=0.95\columnwidth]{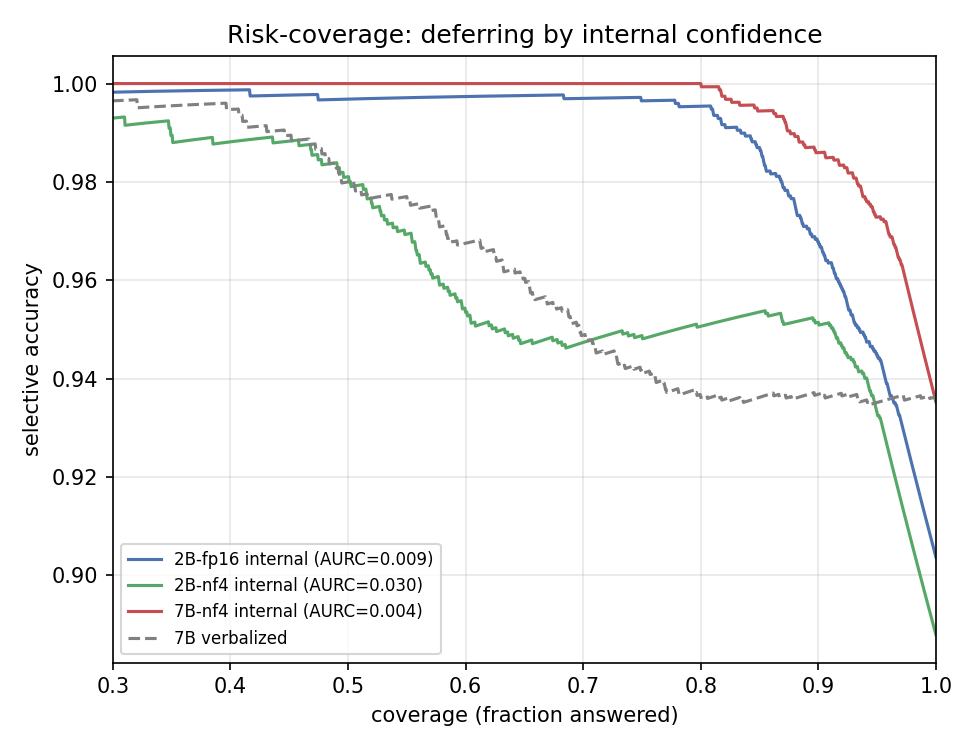}
\caption{Risk--coverage by internal confidence. 7B-4bit (red) answers the most
confident $\sim$80\% of inputs at essentially 100\% selective accuracy; deferring
by 7B \emph{verbalized} confidence (grey dashed) is far weaker. AURC values in the
legend are computed over the pooled predictions; Table~\ref{tab:means} reports the
mean of the per-condition AURC, which is slightly larger.}
\label{fig:rc}
\end{figure}

The three arms are the three configurations that fit a 16\,GB GPU. 7B-4bit
dominates: it has the highest accuracy (0.935), the highest internal AUROC (0.98),
and the lowest area under the risk--coverage curve. Notably it beats even the
full-precision small model on the confidence signal (internal AUROC 0.98 vs 0.95,
AURC 0.008 vs 0.017), so scale's benefit to the uncertainty signal more than
offsets quantization's harm. Figure~\ref{fig:rc} shows the deferral consequence:
ranking predictions by 7B internal confidence, the model answers roughly the most
confident 80\% of inputs at essentially 100\% selective accuracy before any
degradation appears in the retained set. The 7B \emph{verbalized} signal, by
contrast, gives a much flatter, less useful risk--coverage curve on the same
predictions --- deferring by stated confidence is close to deferring at random.
The practitioner's recommendation is therefore unambiguous: spend the memory
budget on parameters, not precision.

\paragraph{Threshold transfer.}
A deployed system fixes a single deferral threshold and cannot retune it per
condition, so what matters is whether a threshold calibrated on clean images stays
safe under degradation. We fit an internal-confidence threshold on clean images to
answer 90\% of them, then apply it unchanged to every condition
(Table~\ref{tab:transfer}). For 7B-4bit the threshold transfers safely: selective
accuracy stays at $\approx\!1.00$ in every degraded condition and \emph{no} errors
are admitted above threshold; the cost of degradation is paid entirely in coverage,
which falls gracefully to 0.39 at low light s3 as the model correctly becomes less
confident. For 2B-4bit the same clean-calibrated threshold \emph{silently admits
errors}: selective accuracy falls to 0.82 at glare s3 with 12\% of all inputs
answered wrongly above threshold. The difference is a direct consequence of the
scale gap in internal AUROC (0.80 vs 0.98): only the larger model's confidence is
discriminative enough that a fixed threshold keeps its promise under shift. This is
the operational form of the deployment recommendation --- the larger quantized
model is not just more accurate, it is the only one whose deferral threshold can be
trusted after calibration.

\begin{table}[t]
\centering
\small
\begin{tabular}{@{}llccc@{}}
\toprule
Arm & Condition & Cov. & Sel.\ acc. & Err.\ adm. \\
\midrule
\multirow{3}{*}{2B 4-bit}
 & clean        & 0.90 & 0.98 & 2\% \\
 & low light s3 & 0.48 & 0.88 & 6\% \\
 & glare s3     & 0.65 & 0.82 & 12\% \\
\midrule
\multirow{3}{*}{7B 4-bit}
 & clean        & 0.90 & 1.00 & 0\% \\
 & low light s3 & 0.39 & 1.00 & 0\% \\
 & glare s3     & 0.44 & 1.00 & 0\% \\
\bottomrule
\end{tabular}
\caption{Threshold transfer. An internal-confidence threshold fit on clean images
to answer 90\% of them, applied unchanged to degraded conditions. Coverage =
fraction answered; Sel.\ acc.\ = accuracy among answered; Err.\ adm.\ = fraction of
all inputs answered incorrectly above threshold. The 7B threshold never admits
errors (it defers instead); the 2B-4bit threshold leaks errors under degradation.}
\label{tab:transfer}
\end{table}

\subsection{Calibration error would hide this}
Expected Calibration Error is low for every arm and every signal (0.01--0.07),
because all arms are accurate and confident on most conditions. The clearest
illustration is arm A: its \emph{verbalized} confidence has an ECE of just
$0.011$ --- better ``calibrated'' than its own internal signal (ECE $0.053$) ---
yet the same verbalized confidence separates correct from incorrect answers at
AUROC $0.56$, barely above chance, while the internal signal reaches $0.95$ on the
identical predictions. A confidence value that is nearly constant at $\approx\!0.9$
on a 90\%-accurate condition is well calibrated and completely useless for
ranking. Reliability diagrams and ECE assess whether confidence \emph{magnitude}
matches accuracy; error-detection AUROC assesses whether confidence
\emph{discriminates} correct from incorrect answers. ECE does not distinguish the
arms; AUROC does, which is why we adopt it as the primary metric and recommend it
for future work on VLM confidence.

% ===========================================================================
\section{Discussion}
% ===========================================================================
Two levers a practitioner controls --- parameters and precision --- act on the
confidence signal in opposite directions and through different channels. Scale
mostly improves the \emph{internal} signal, lifting error-detection AUROC to near
ceiling, while barely improving the model's ability to \emph{state} a
discriminative confidence. Quantization mostly damages the signal --- both the
internal AUROC and the model's willingness to emit a confidence number --- while
leaving accuracy almost untouched. The two effects are not symmetric: the scale
gain (internal AUROC $0.80\!\to\!0.98$) exceeds the quantization loss
($0.95\!\to\!0.80$), which is why the larger quantized model wins overall.

The most striking result is that verbalized confidence does not become a usable
error detector with scale. A tempting reading of a single small-model study would
be that stated confidence is poor because the model is small; our data contradict
that. The 7B model states a confidence number every time and is far more accurate,
yet its stated numbers separate right from wrong answers barely above chance in a
third of conditions, while its own token probabilities do so almost perfectly on
the identical predictions. Verbalized confidence appears to be a learned surface
behaviour that scale makes more \emph{fluent} (always parseable) without making it
more \emph{truthful}, whereas internal probability is a direct read of the model's
distribution that scale sharpens. The practical implication is that stated
confidence should not be used as a deferral signal for these models regardless of
size; mean token probability should.

For deployment, the recommendation is concrete. Under a fixed memory budget,
prefer a larger quantized model to a smaller full-precision one, and defer using
internal token probability rather than stated confidence. But no configuration is
safe under severe underexposure: at low light s3 every arm loses accuracy sharply
(down to 0.80 even at 7B) and the internal signal weakens (AUROC 0.89 at 7B, with
a bootstrap interval reaching 0.80), so a system that may encounter severe low
light needs an explicit upstream image-quality check, not a downstream confidence
threshold alone.

% ===========================================================================
\section{Limitations}
% ===========================================================================
Two scale points give a direction, not a scaling law, and cannot locate an
emergence threshold. Quantization is measured at one model size, bounding but not
eliminating its confound with scale. We use $n\approx100$ per condition, so some
bootstrap intervals are wide and conditions near ceiling accuracy yield undefined
AUROC. All results are on Food101 posed as four-way multiple choice; generality
to open-ended VQA is untested. We study a single model family, so scale effects
may be family-specific. The multiple-choice format forces an answer and therefore
cannot observe refusal behaviour. Finally, our degradations approximate rather
than reproduce real camera artifacts.

% ===========================================================================
\section{Conclusion}
% ===========================================================================
We measured how model scale and 4-bit quantization affect the two available
confidence signals in the Qwen2-VL family under realistic image degradation,
across 5{,}700 predictions and three configurations that each fit a single 16\,GB
GPU. Scale sharply improves the model's internal error-detection signal (mean
AUROC $0.80\!\to\!0.98$) but leaves its verbalized confidence weak and often at
chance, so the gap between what the model knows and what it says widens with size.
Quantization is nearly free for accuracy yet costly for the confidence signal.
For a fixed memory budget the larger quantized model is the best choice on both
accuracy and uncertainty, and internal token probability, not stated confidence,
is the signal to defer on --- except under severe low light, where no
configuration is reliable and an upstream image-quality check is required.

% ===========================================================================
\section*{Reproducibility}
% ===========================================================================
All code, the experiment notebook, per-prediction CSVs, and figures are released
at \url{https://github.com/Asif-Ferdous/vlm-scale-quant}. Every arm reproduces on
a single free-tier NVIDIA T4; random seeds are fixed throughout, and an item
manifest enforces identical inputs across arms.

\appendix
% ===========================================================================
\section{Elicitation templates and parse rates}
\label{app:templates}
% ===========================================================================
Three templates were piloted on 20 clean items per model size. The few-shot
template (a worked two-line example) was selected for the main runs. Parse rates
at 7B were 100\% (few-shot), 100\% (direct) and 55\% (minimal); the two selected
formats agree on 100\% accuracy, so the finding does not rest on a single prompt.
At 2B the same few-shot template yields a 64\% parse rate averaged over all
conditions, versus 100\% at 7B --- the scale effect on format-following reported
in Section~\ref{sec:results}. Templates are reproduced verbatim below; {question}
and {options} are substituted at runtime.

\paragraph{Few-shot (selected).}
\begin{quote}\small\ttfamily
You must reply in exactly two lines.\\[2pt]
Example reply:\\
Answer: pizza\\
Confidence: 85\\[2pt]
Now do the same for this image.\\
Question: \{question\}\\
Options: \{options\}
\end{quote}

\paragraph{Direct.}
\begin{quote}\small\ttfamily
Question: \{question\}\\
Options: \{options\}\\
Answer the question, then rate your confidence from 0 to 100.\\
Use exactly this format and nothing else:\\
Answer: <option>\\
Confidence: <number>
\end{quote}

\paragraph{Minimal.}
\begin{quote}\small\ttfamily
\{question\}\\
Choose one: \{options\}\\[2pt]
Reply with two lines only:\\
Answer: (your choice)\\
Confidence: (0-100)
\end{quote}

\paragraph{Parsing.} Answers were extracted with \texttt{Answer:\textbackslash
s*(.+)} and confidences with \texttt{Confidence:\textbackslash s*([0-9]\{1,3\})},
the latter divided by 100. Replies with no explicit answer line (a bare label)
were scored against the whole reply, so a correct bare answer is not penalised for
skipping the format; the internal confidence for such replies is the mean token
probability over the full generated span.

\bibliography{references}
\bibliographystyle{plainnat}

\end{document}